\documentclass[submission,copyright,creativecommons]{eptcs}
\usepackage{amsmath}
\usepackage{graphicx}
\usepackage{multirow}
\usepackage{listings}
\usepackage{svg}
\usepackage{url}
\usepackage{array}

\providecommand{\event}{International Conference on Logic Programming 2023 (ICLP’23).} 

\usepackage{iftex}

\ifpdf
  \usepackage{underscore}         
  \usepackage[T1]{fontenc}        
\else
  \usepackage{breakurl}           
\fi

\title{Explainable and Trustworthy Traffic Sign Detection for Safe Autonomous Driving: \\ An Inductive Logic Programming Approach}
\author{Zahra Chaghazardi
\institute{Department of Computer Science, University of Surrey\\
United Kingdom}
\email{z.chaghazardi@surrey.ac.uk}
\and
Saber Fallah 
\institute{Connected and Autonomous Vehicles Lab, University of Surrey\\
United Kingdom}
\email{s.fallah@surrey.ac.uk}
\and
Alireza Tamaddoni-Nezhad
\institute{Department of Computer Science, University of Surrey\\
United Kingdom}
\email{a.tamaddoni-nezhad@surrey.ac.uk}
}

\def\titlerunning{Traffic Sign Detection for Safe Autonomous Driving}
\def\authorrunning{Z.  Chaghazardi, S. Fallah \& A. Tamaddoni-Nezhad}

\begin{document}
\maketitle

\begin{abstract}
Traffic sign detection is a critical task in the operation of Autonomous Vehicles (AV), as it ensures the safety of all road users. Current DNN-based sign classification systems rely on pixel-level features to detect traffic signs and can be susceptible to adversarial attacks. These attacks involve small, imperceptible changes to a sign that can cause traditional classifiers to misidentify the sign. We propose an Inductive Logic Programming (ILP) based approach for stop sign detection in AVs to address this issue. This method utilises high-level features of a sign, such as its shape, colour, and text, to detect categories of traffic signs. This approach is more robust against adversarial attacks, as it mimics human-like perception and is less susceptible to the limitations of current DNN classifiers.
We consider two adversarial attacking methods to evaluate our approach: Robust Physical Perturbation (PR2) and Adversarial Camouflage (AdvCam). These attacks are able to deceive DNN classifiers, causing them to misidentify stop signs as other signs with high confidence. The results show that the proposed ILP-based technique is able to correctly identify all targeted stop signs, even in the presence of PR2 and ADvCam attacks. The proposed learning method is also efficient as it requires minimal training data. Moreover, it is fully explainable, making it possible to debug AVs.
\end{abstract}

\section{Introduction}

The popularity of AVs is rising rapidly because of their potential to reduce human error on the road, leading to safer transportation. AVs are believed to make more accurate perceptions and react faster than humans. 
Deep Neural Networks (DNNs) play a significant role in developing perception systems for AVs. However, DNNs face significant challenges that must be addressed before AVs can be deployed safely \cite{chaghazardi2023logic}. The major challenges facing DNN-based vision systems in autonomous driving are discussed below.

DNN-based systems are often considered "black boxes" because their logic is not transparent. Since it is difficult to explain how the system makes the prediction, it is challenging to debug them when they make a wrong decision. For example, misclassifying objects, such as mistaking shadows for pedestrians, is a common problem in AVs and making decisions based on these misclassifications can lead to fatal accidents. Considering the fatal Uber accident \cite{uber}, given that the AV's DNN-based decision-making is opaque, there is no way to debug the system and ensure such mistakes do not happen again. Moreover, using algorithms with ambiguous logic makes it impossible to evaluate and trust them. This means that regulatory approval is not applicable to stochastic-based AV vehicles.

Furthermore, DNNs face significant challenges when it comes to learning from small data and achieving out-of-distribution generalizability and transferability to new domains. In real-world scenarios, particularly in security domains, there is often a lack of large, annotated, and carefully curated data sets to train these systems. This can make it difficult for DNNs to acquire knowledge from a few examples and transfer it to new domains, unlike humans, who can do so with ease. Anomaly detection tasks, in particular, are affected by this challenge due to the rarity of anomalous data. Anomalies can be caused by errors, faults, or adversarial attacks, which can lead to security and safety hazards. Adversarial examples provide evidence of a network's weakness in achieving high generalisation performance \cite{szegedy2013intriguing}. Improving generalizability is crucial for adapting models to new domains when there is insufficient data. Given the lack of generalizability, current DNNs are not able to incrementally learn and improve when deployed in real-life situations and transfer knowledge from one domain to another (multi-domain) \cite {salami2021state}.

In the real world, DNNs are vulnerable to adversarial attacks and can be deceived easily. In adversarial cases, minor perturbations will lead to misclassifications with high confidence. Adversarial attacks have been investigated for different vision tasks, such as image classification, object detection, and semantic segmentation. For example, it is possible to change the red traffic light to green for AV \cite{yan2022rolling}, make people invisible to AI \cite{thys2019fooling} using small crafted adversarial patches held in front of the body or make the AV to misinterpret a stop sign as a speed limit sign \cite{eykholt2018robust}.

Researchers have suggested a few solutions, such as transfer learning for transferring knowledge to another domain, to address challenges associated with DNN classifiers. However, the proposed solutions partially solve the problems and have many limitations. For example, the transfer learning approach faces a significant challenge regarding data sharing and several legal issues such as privacy and property law \cite{kop2020machine}.  

To strengthen the safety of autonomous driving, this paper proposes an explainable ILP-based solution focusing on traffic sign detection. The proposed method mimics human perception to recognise traffic signs by detecting high-level features, including signs' geometric shapes, colours and contents, that differentiate them from other signs. While DNNs only use low-level (pixel-level) features that can be easily misled \cite{eykholt2018robust} and need a large amount of data, this traffic sign detector only needs a handful of training images and is fully robust against adversarial attacks.

Several studies have investigated the application of Inductive Logic Programming (ILP) in image recognition tasks. ILP has been employed in Logical Vision \cite{dai2015logical, dai2018logical}, incorporating the abductive perception technique \cite{shanahan2005perception} to extract high-level interpretation of objects such as classical 2D shapes by utilising low-level primitives, such as high contrast points. ILP has also been used for 3D scene analysis \cite{farid2014plane} with 3D point cloud data. However, to our knowledge, a traffic sign detection based on the ILP has not been proposed previously for traffic sign classification. Therefore, our approach is a novel contribution to this context.

The paper is structured as follows. Section 2 surveys some successful adversarial examples in AVs. Section 3 describes the framework for robust traffic sign detection using ILP. Section 4 details experiments. In this section, the Aleph-based approach is compared with the Metagol-based approach. Metagol can learn hypotheses with only one positive and one negative example, while Aleph needs at least eight positive and negative examples to have the same accuracy as Metagol. Also, the ILP-based system is compared with the DNN-based classifier on adversarial examples. The results show that the ILP-based approach is considerably more resilient to adversarial attacks. Finally, Section 5 summarises the outcomes and discusses further work.

\section{Adversarial Attacks on AVs' Perception}

In this section, we survey a sample of successful adversarial attacks in autonomous driving that easily deceived DNN-based vision classifiers. An adversarial attack aims to generate adversarial examples as the input for machine learning systems.  However, adversarial examples are only negligibly modified from the real examples; they lead to misclassification \cite{gui2021review}.

When the fragility of deep neural networks to specific input perturbations was discovered for the first time, it was shown that an adversarial attack could turn a bus into an ostrich for an AI system \cite{szegedy2013intriguing}. Another algorithm named Show-and-Fool \cite{chen2017attacking} was introduced to evaluate the robustness of an image captioning system. This method attained a 95.8\% attack success rate for adversarial examples via applying a minor perturbation on image pixels which are invisible to humans, turning a stop sign into a teddy bear for the AI system.

The authors of \cite{hendrik2017universal}  devised a method whereby semantic image segmentation could be attacked using adversarial perturbation to blend out (vanish) a desired target. They showed the existence of universal noise, which removes a target class (e.g. all pedestrians) from the segmentation while leaving it mostly unchanged otherwise.
The robustness of the popular DNN-based semantic segmentation models  evaluated against adversarial attacks on urban scene segmentation \cite{arnab2018robustness}. The results showed that the segmentation performances of all models seriously dropped after the attacks.

 Later it was shown that adversarial examples could be misclassified by deep learning systems in real life \cite{kurakin2018adversarial}. Previous works have threatened the model by feeding machine learning classifiers directly, which is not always possible in the real world.

Another paper \cite{eykholt2018robust} proposed the Robust Physical Perturbations (RP2) technique to fool a Convolutional Neural Network (CNN) based road sign classifier in the physical world under various distances and viewpoints using different robust visual adversarial perturbations. This approach caused targeted misclassification, which changed a stop sign into a speed limit sign for the AI system. They also proposed a disappearance attack, causing a stop sign hidden from state-of-art object detectors like Mask R-CNN and YOLO \cite{eykholt2018physical}. 
An Adversarial Camouflage (AdvCam) approach \cite{duan2020adversarial} generated adversarial photos to fool a DNN classifier at various detecting angles and distances. With a few stains invisible to humans, this technique can cause the classifier to misclassify the objects, such as misidentifying a stop sign as a "barber shop" with .82\% confidence.

\begin{figure}[ht]
    \centering
    \includegraphics[width=\linewidth]{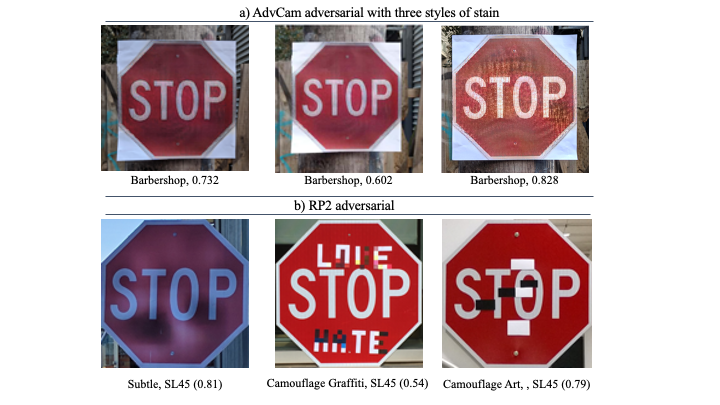}
    \caption{Targeted physical perturbation by a) AdvCam and b) RP\_2 misleading DNN classifiers, SL45 is speed limit 45 sign. }
    \label{fig:attacks}
\end{figure}

Fig. \ref{fig:attacks} illustrates targeted stop signs with successful physical-world attacking approaches named RP2 and AdvCam, misleading the state-of-the-art DNN classifiers.

An Adaptive Square Attack (ASA) method \cite{li2020adaptive} has been suggested that can attack the black box by generating invisible perturbation for traffic sign images, successfully leading to sign misclassification. Five adversarial attacks and four defence methods have been investigated on three driving models adopted in modern AVs \cite{deng2020analysis}. They demonstrated that while these defence methods can effectively defend against a variety of attacks, none can provide adequate protection against all five attacks.

One recent work proposed three sticker application methods, namely RSA, SSA and MCSA, that can deceive the traffic sign recognition DNNs with realistic-looking stickers \cite{bayzidi2022traffic}.
Another attack included painting the road, which targeted deep neural network models for end-to-end autonomous driving control  \cite{boloor2019simple}. Another work demonstrated a successful physical adversarial attack on a commercial classification system to deceive an AV's sign classifier\cite{morgulis2019fooling}.

BadNets algorithm \cite{gu2019badnets} was implemented to deceive a complex traffic sign detection system leading to maliciously misclassifying stop signs as speed-limit signs on real-world images.
 
These adversarial attacks on the deep-learning models pose a significant security threat to autonomous driving. 

\section{Robust Traffic Sign Detection Using ILP}
Inductive Logic Programming (ILP) is a machine learning method which uses logic-based representation and inference. 
Depending on the type of logical inference and the search algorithm, there are different ILP systems, such as Aleph \cite{Srinivasan2001} and Metagol \cite{metagol}, that used in this paper.

Due to a logic-based representation and inference, ILP has the potential for human-like abstraction and reasoning. These logic-based AI approaches have the ability to learn unknown complex tasks with only a few examples. It complements deep learning because logic programs are interpretable and data-efficient, leading them towards a strong generalisation. Moreover, these rule-based approaches, which are explicitly symbolic, are sometimes considered safer than neural approaches \cite{anderson2020neurosymbolic}. 

ILP aims to learn a hypothesis (rule) using a few positive and negative examples and Background Knowledge (BK); this induced rule, alongside BK, should cover as many positive and as few negative examples as possible \cite{muggleton1991inductive}. For inducing the rules, BK should include all essential predicates to represent the relevant information.

One of the advantages of ILP is its ability to use BK, including facts and rules in the form of logical expressions, which could be related. In ILP, choosing appropriate BK based on well-selected features is essential to obtaining good results \cite{cropper2020turning}. Moreover, using BK makes ILP incremental. For example, suppose we want to learn animal signs in traffic sign detection, choosing sign "a" contains an animal( {\em contains(a, animal)} as a BK, which holds when traffic sign "a" has an animal symbol. Then we provide BK with various different animal shaped symbols (deer, cow, …). In that case, if we see a new animal sign that doesn't exist in our BK, we can add it to our BK without relearning, and there is no need to change the hypothesis. This feature makes it possible to have real-time interaction with drivers towards customised autonomous driving.

\begin{figure}[ht]
    \centering
    \includegraphics[width=.8\linewidth]{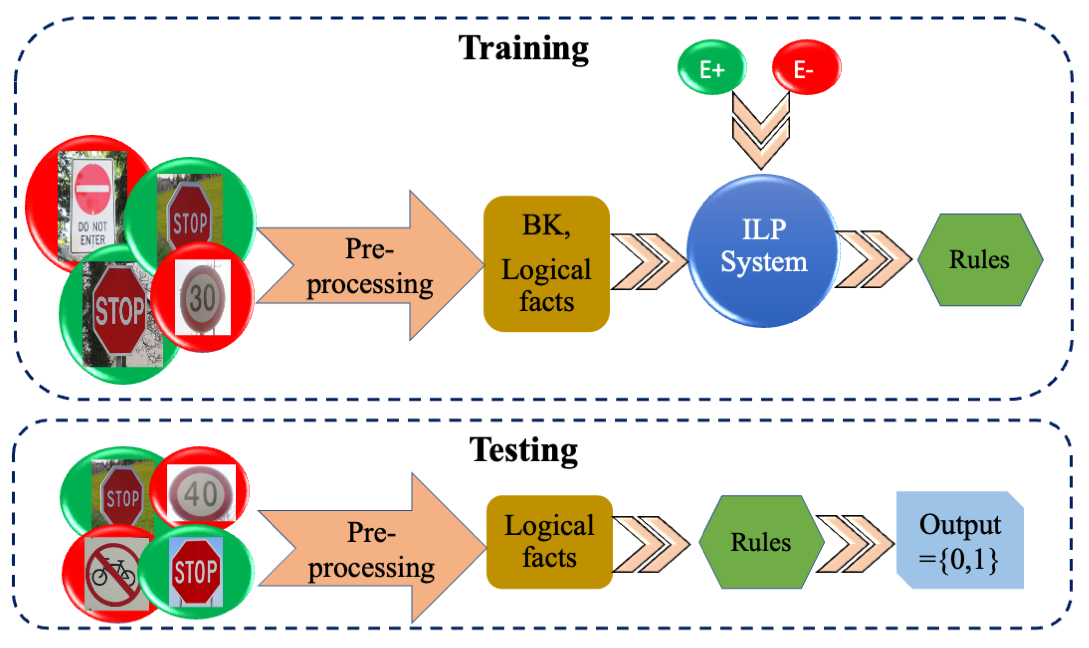}
    \caption{ILP- based traffic sign classifier}
    \label{fig:diagram}
\end{figure}

Our proposed ILP-based stop sign detection system is demonstrated in  Fig. \ref{fig:diagram}. The first step is pre-processing all the images, including training and test images, and turning them into a symbolic representation to provide BK. In the pre-processing phase, high-level features of traffic sign images, including colour, shape, text and digits, are extracted and represented as a set of logical facts for the next step. For feature extraction, computer vision tools such as OpenCV can extract high-level features using low-level features such as pixel colours or colour gradients.

In the next step, a set of positive and negative training examples (E) and a set of logical facts as BK extracted from the previous step will be provided to the ILP system. The system aims to learn a hypothesis H such that $B, H\models E$ where $\models$ is logical entailment.   

We use Aleph and Metagol as the ILP system to induce the rule for stop sign detection.

We used Aleph5 as the ILP system in one of our experiments; it is an old ILP system developed in Prolog and based on inverse entailment. Aleph's algorithm resolves the relationship between the determination predicate and the determining predicate to generate a general theory.

Metagol is employed in our other experiment. It is an ILP system based on Meta Interpretive Learning (MIL) \cite{muggleton2015meta} implemented in Prolog. By instantiating metarules, MIL learns logic programs from examples and BK. In addition, MIL not only learns the recursive definition and fetches higher-order meta-rules but also supports predicate invention. 

\section{Experimental Evaluation}
This experiment aims to learn {\em "traffic\_sign"}, which is the target predicate. For simplicity, only the stop sign is investigated; other traffic signs can be included to have a complete traffic sign classifier. 
We provide Aleph and Metagol with the same BK. The Aleph  mode declarations are illustrated in  Table \ref{tab:Modes}, and the Metagol-based system is supplied with the metarules demonstrated in Table \ref{tab:metarules}, uppercase letters represent predicate symbols (second-order variables), and lowercase letters represent variables.

\begin{table}[h!]
 \centering
 \caption{Mode declarations for Aleph experiments.}
 \label{tab:Modes}
 {\begin{tabular}{@{\extracolsep{\fill}}l}
   \hline
       $:- modeh ( 1 , traffic\_sign(+sign, \#class).$\\ 
       $:- modeb ( *, colour(+sign,\#colour)).$\\
       $:- modeb ( *, shape(+sign,\#shape)).$\\
       $:- modeb ( *, word(+sign,-w)).$\\
       $:- modeb ( *, closely\_match(+w, \#word)).$\\
       $:- modeb ( *, number(+sign, -n)).$\\
       $:- modeb ( *, digits(+n,\#int)).$\\
   \hline
    \end{tabular}}
\end{table}

\begin{table}
    \centering
     \caption{Employed metarules in Metagol experiment.  }
    \label{tab:metarules}
    {\begin{tabular}{@{\extracolsep{\fill}}ll}
        \hline
        Name  & Metarule  \\
        \hline
        Identify     & P(x,y)$\longleftarrow$  Q(x,y)              \\
        Inverse      & P(x,y)$\longleftarrow$  Q(y,x)              \\
        Precon       & P(x, y)$\longleftarrow$ Q(x), R(x, y)       \\
        Postcon      & P(x, y)$\longleftarrow$ Q(x, y), R(y)       \\
        Chain        & P(x, y)$\longleftarrow$ Q(x, z), R(z, y)    \\
        Recursion    & P(x, y)$\longleftarrow$ Q(x, z), P (z, y)   \\
        \hline
    \end{tabular}}
\end{table}

In Aleph mode declaration, {\em "modeh"} indicates that the predicate should appear in the head of the hypothesis, and {\em "modeb"} indicates that it should be in the body of the induced hypothesis. According to Table \ref{tab:Modes}, six predicates can be used in the body of the induced hypothesis which. The meaning of each predicate is defined as follows:

\begin{itemize}
    \item {\em traffic\_sign(a, \#class)}, which holds when the sign "a" belongs to a specific category of traffic sign determined by \#class (e.g. a stop sign).
    \item {\em colour(a, \#colour)}, which holds when a certain \#colour(e.g. red) exists in the sign "a".
    \item {\em shape(a, \#shape)}, which holds when the shape of sign "a" is a specific shape determined by \# shape(e.g. circle).
    \item {\em has\_word(a, a\_w1)}, which holds when the sign "a" has the word a\_w1 on it.
    \item  {\em closely\_match(a\_w1, w)}, which holds when the word "a\_w1" closely matches the word "w" (e.g. stop).
    \item {\em number(a, a\_n1)}, which holds when the sign "a" has the number "a\_n1".
    \item  {\em digits(a\_n1,n)}, which holds when the number "a\_n1" includes "n" (e.g. 60)
\end{itemize}

To further explain the process of converting images into a set of logical facts,
we take one positive example, a stop sign named p1, and one negative example, a speed limit sign named n1. In the pre-processing stage, the high-level features of these traffic signs were extracted to be included in the BK. The details of these features and their corresponding logical representation are presented in Table \ref{tab:Features}. 

\begin{table}
    \centering
    \caption{Extracted features for a positive (p1) and negative (n1) examples.  }
    \label{tab:Features}
   {\begin{tabular}{ll}
        \hline
        Pos example(p1) & Neg example(n1)  \\
        \hline
        $color\left(p1, red\right).$                  & $color\left(n1,red\right).$\\
        $color\left(p1, white\right).$                & $color\left(n1,white\right).$\\
        $shape\left(p1, octagon\right).$              & $shape\left(n1,Circle\right).$\\
        $has\_word\left(p1, p1\_w1\right).$                & $number\left(n1,n1\_d1\right).$\\
        $closely\_match\left(p1\_w1, stop\right).$    & $digits\left(n1\_d1,30\right).$\\
        \hline
    \end{tabular}}
\end{table}

These logical facts, together with the names of the positive and negative examples (such as sign(p1) as a positive and sign(n1) as a negative example), will enable the ILP system to induce a hypothesis (logical rule). Finally, the ILP system recognises the new traffic signs using this induced rule.

\subsection{Material and Method}

\begin{figure}[ht]
    \centering
    \includegraphics[width=.8\linewidth]{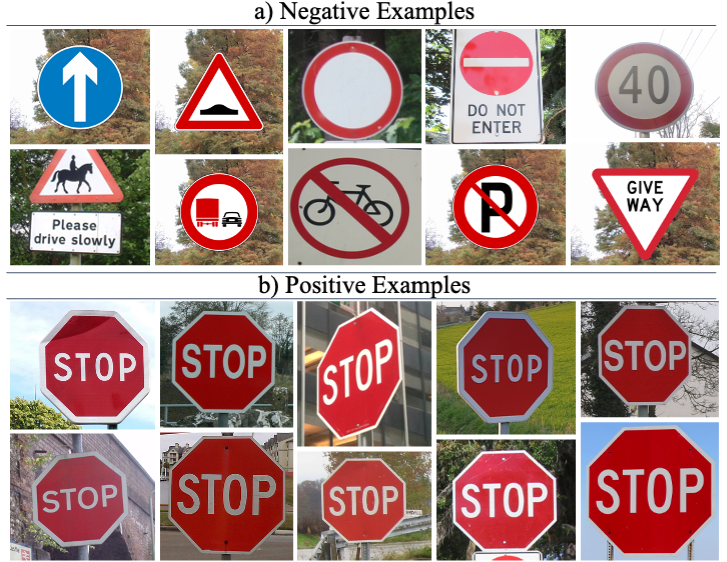}

    \caption{Base data set for training and testing}
    \label{fig:normal}
\end{figure}

\paragraph{Base data set.} Our base data set includes traffic sign images without any adversarial perturbation. They have been downloaded from Wikimedia Commons as no-restriction images. Positive examples contain stop sign images, and negative examples include other traffic signs excluding stop sign images. 
Normal positive and negative examples are shown in Fig. \ref{fig:normal}. 

\paragraph{Adversarial test data set.}To evaluate the robustness of the ILP stop sign detector against adversarial attacks, we used the targetted traffic signs attacked by RP\_2 and AdvCam.  

 RP\_2 is a general attack algorithm for misleading standard-architecture road sign classifiers. It generates visual adversarial perturbations, such as black and white stickers attached to a traffic sign to mislead the classifier. The RP\_2 data set contains three types of perturbation, stop signs perturbed by subtle, camouflage graffiti and camouflage art attacks viewed from different angles.

AdvCam is an approach for creating camouflage physical adversarial images to fool state-of-the-art DNN-based image classifiers. This approach can make the classifier identify a stop sign as a "barber shop" with high confidence. This paper will utilise targetted stop signs with Advcam with different stain styles to evaluate the proposed ILP sign classifier.

The feature recognition element should extract high-level features, including the traffic sign's border shape, colour and text; symbol extraction can be added in future; {\em OpenCV} is employed for this purpose. First, the image background is removed utilising {\em rembg} and then pre-processed by {\em cv2.bilateralFilter} for noise reduction. 

Colour masks were then employed using cv2.inRange for colour detection, and small areas were ignored. After that, by applying morphological operations, colour masks were post-processed.

 For text and digit detection, {\em EasyOCR} is utilised; if the detected item is a word, it will be investigated if the detected word closely matches some common words in traffic signs; for example, it should have at least three letters in common with the word STOP to recognise as a stop word. 
 
For shape detection, {\em cv2.findContours} is applied on detected colour masks, and {\em cv2.approxPolyDP} is utilised for polygon detection. 


  \paragraph{Convolutional Neural Network.}To compare the results, a well-known CNN classifier \cite{Yadav} is utilised that is trained on the German Traffic Sign Recognition Benchmark (GTSRB) \cite{stallkamp2012man}. The evaluation of this architecture achieved 97.6\% accuracy on the GTSRB test data set.

The base data set is utilised for training the ILP systems (Aleph and Metagol). First, we randomly select an equal number of positive and negative examples in each run, so the default accuracy is 50\% for this training data set. Next, the ILP-based systems try to find a hypothesis that covers as many positive and as few negative examples as possible. Then the remaining examples in the data set are used as a test data set for evaluation to determine the accuracy. This process is repeated one hundred times, and average accuracy is calculated for each certain number of positive and negative examples in the training set.  Therefore we have a fair comparison between Aleph and Metagol regarding the size of the required training data set.

The data and the code used in this experiment are available on GitHub \cite {sign}.

\subsection{Results and Discussion}

\begin{figure}[ht]
    \centering
    \includegraphics[width=.8\linewidth]{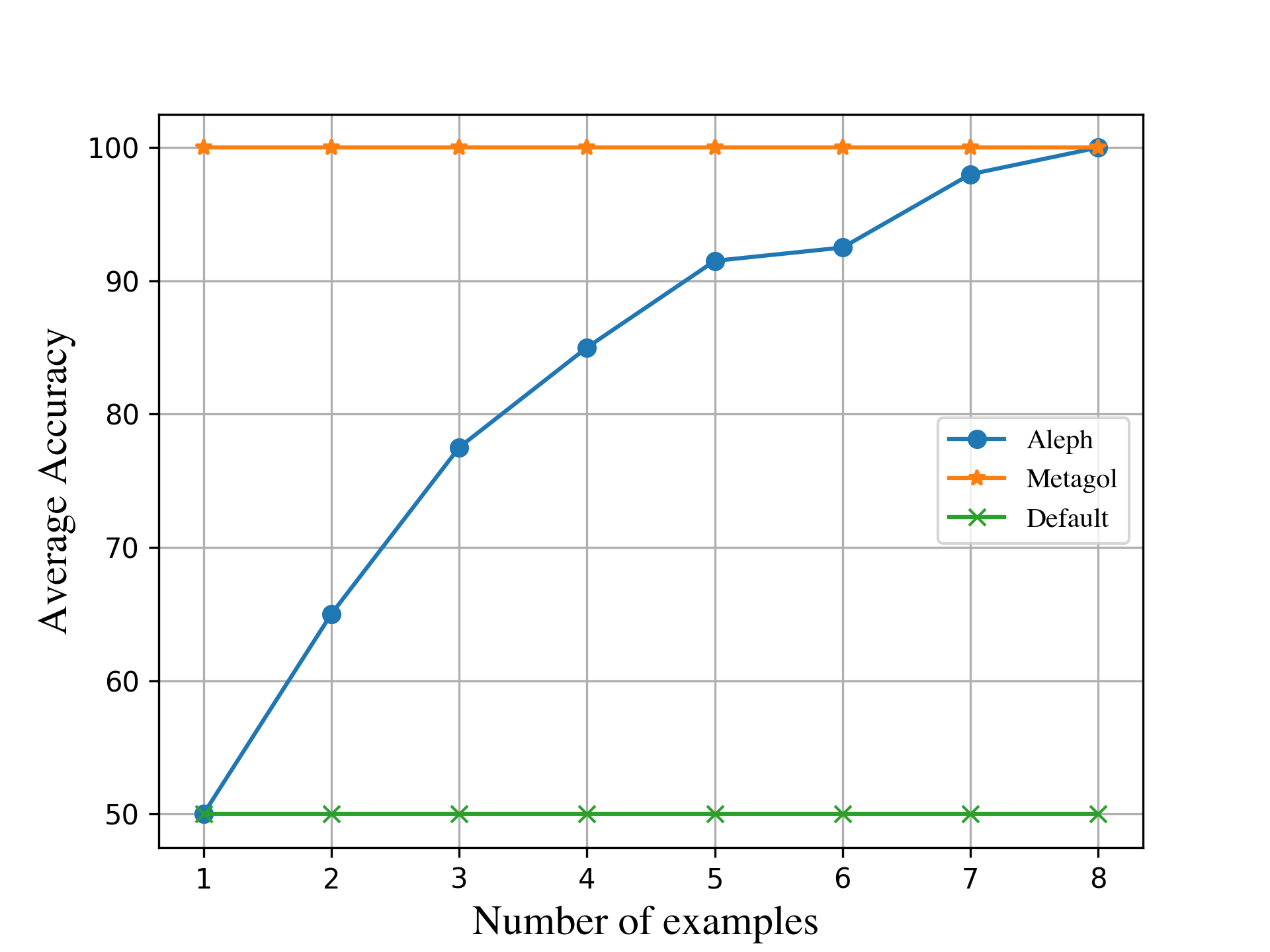}

    \caption{ Average accuracy of Aleph vs Metagol with an increasing number of training examples from the base data set (equal positive and negative sets) }
    \label{fig:comparision}
\end{figure}

Fig. \ref{fig:comparision} illustrates the average accuracy of Aleph and Metagol-based ILP systems with increasing training examples. According to this figure, Metagol can find a hypothesis with 100\% accuracy on the test data set including only one positive and one negative example. In comparison, Aleph starts learning with at least two positive and two negative examples with around 65\% accuracy. Aleph can reach the same level of accuracy as Metagol by learning from eight positive and negative examples. According to these results, Metagal is more data-efficient than Aleph. In this figure, the orange curve shows the default accuracy, which is equal to 50\% because the number of negative and positive examples are equal in each run.  

The hypothesis (a logic program) induced by Metagol with only one set of positive and negative examples is the same as the learned rule by Aleph with eight positive and negative examples. It is entirely accurate on the base test data set and is shown below:



\begin{minipage}{\linewidth}
\begin{lstlisting}
traffic_sign(A,stop_sign):-
      has_word(A,A_w1), 
      closely_match(A_w1,stop).
\end{lstlisting}
\end{minipage}
       
This learned rule is completely explainable and matches human interpretation. The rule says the traffic sign {\em "A"} is a stop sign when the two literals {\em has\_word(A, A\_w1)} and  {\em closely\_match(A\_w1, stop)} hold, i.e. if the sign contains a word and that word closely matches stop, that sign would be predicted a stop sign.

The performance of this hypothesis is evaluated on the base data set and attack data sets, including RP\_2 (subtle, camouflage graffiti and camouflage art attacks) and AdvCam with different stains. The accuracy of this rule on all  test data sets is 100\%. While the ILP-based sign detector shows a perfect performance, the DNN-based classifier shows abysmal performance facing manipulated images.

Table \ref{tab:results} compares the results of the DNN-based classifier and the ILP-based classifier on different data sets. The DNN-based classifier is trained on the GTSRB data set, which contains more than 50,000 images. The Aleph-based classifier is trained on eight positive and negative examples, while the Metagol approach is trained on only one negative and one positive example. It shows that while ILP-based systems can learn from small amounts of data, they are more resilient to noise and adversarial attacks.

\begin{table}
   \centering
   \begin{tabular}{l l|c|c}
       \hline
       \multicolumn{2}{c|}{data set} & DNN-based  & ILP-based  \\
       \hline
        \multicolumn{2}{c|}{Base}                   &  100\%  &  100\%     \\
        \hline
       \multirow{3}{*}{RP\_2} & subtle                   &  0  &  100\%       \\
       & camouflage graffiti      &  0\%  &  100\%     \\
       & camouflage art attacks   &  6.6\%  &  100\%     \\
        \hline       
       \multicolumn{2}{c|}{AdvCam}                  &  66.6\%  &  100\%     \\

       \hline
   \end{tabular}
   \caption{Comparing the results of the hypothesis induced by the proposed ILP-based approach (Aleph and Metagol) on different test data sets with a DNN classifier.}
   \label{tab:results}
\end{table}

\section{Conclusions}

DNN-based traffic sign classifiers need a large amount of data for training, and it has been shown that they are vulnerable to adversarial attacks or natural noise. They are also not explainable;  consequently, there is no way to debug them. While DNN-based classifiers suffer from these problems, we propose an ILP-based approach for traffic sign detection in autonomous vehicles to address these issues.

Our proposed technique mimics humans in traffic sign detection and uses high-level features of a sign, such as colour and shape, for detection. Therefore this method is data efficient, explainable and able to withstand adversarial attacks that cannot easily deceive humans.

The results indicate that our approach with only a handful of training data can induce logical rules easily understandable by humans. Furthermore, it significantly outperforms the deep learning approach regarding adversarial attacks. It shows a 100\% accuracy on the data set targetted with RP\_2 and AdvCam attacking approaches, while a DNN-based classifier performs poorly on these data sets.

For future works, we suggest employing DNN for high-level feature extraction (shapes or symbols) in traffic signs. Integrate Machine Learning and Logic programming in AV applications will use both the strengths of machine learning and symbolic AI (knowledge and reasoning) to address the AI obstacles.


 
\section*{Acknowledgments}
The first author would like to acknowledge her PhD grant funding from the Breaking Barriers Studentship Award at the University of Surrey. Also, we would like to acknowledge the support of Dany Varghese with Aleph PyILP \cite {pyilp}.

\bibliographystyle{eptcs}
\bibliography{EPTC23}

\begin{thebibliography}{10}
\providecommand{\bibitemdeclare}[2]{}
\providecommand{\surnamestart}{}
\providecommand{\surnameend}{}
\providecommand{\urlprefix}{Available at }
\providecommand{\url}[1]{\texttt{#1}}
\providecommand{\href}[2]{\texttt{#2}}
\providecommand{\urlalt}[2]{\href{#1}{#2}}
\providecommand{\doi}[1]{doi:\urlalt{https://doi.org/#1}{#1}}
\providecommand{\eprint}[1]{arXiv:\urlalt{https://arxiv.org/abs/#1}{#1}}
\providecommand{\bibinfo}[2]{#2}

\bibitemdeclare{article}{anderson2020neurosymbolic}
\bibitem{anderson2020neurosymbolic}
\bibinfo{author}{Greg \surnamestart Anderson\surnameend},
  \bibinfo{author}{Abhinav \surnamestart Verma\surnameend},
  \bibinfo{author}{Isil \surnamestart Dillig\surnameend} \&
  \bibinfo{author}{Swarat \surnamestart Chaudhuri\surnameend}
  (\bibinfo{year}{2020}): \emph{\bibinfo{title}{Neurosymbolic reinforcement
  learning with formally verified exploration}}.
\newblock {\slshape \bibinfo{journal}{Advances in neural information processing
  systems}} \bibinfo{volume}{33}, pp. \bibinfo{pages}{6172--6183}.

\bibitemdeclare{inproceedings}{arnab2018robustness}
\bibitem{arnab2018robustness}
\bibinfo{author}{Anurag \surnamestart Arnab\surnameend},
  \bibinfo{author}{Ondrej \surnamestart Miksik\surnameend} \&
  \bibinfo{author}{Philip~HS \surnamestart Torr\surnameend}
  (\bibinfo{year}{2018}): \emph{\bibinfo{title}{On the robustness of semantic
  segmentation models to adversarial attacks}}.
\newblock In: {\slshape \bibinfo{booktitle}{Proceedings of the IEEE Conference
  on Computer Vision and Pattern Recognition}}, pp. \bibinfo{pages}{888--897}.

\bibitemdeclare{misc}{Srinivasan2001}
\bibitem{Srinivasan2001}
\bibinfo{author}{\surnamestart {Ashwin Srinivasan}\surnameend}
  (\bibinfo{year}{2001}): \emph{\bibinfo{title}{The aleph manual}}.
\newblock
  \bibinfo{howpublished}{\url{https://www.cs.ox.ac.uk/activities/programinduction/Aleph/aleph.html}}.

\bibitemdeclare{inproceedings}{bayzidi2022traffic}
\bibitem{bayzidi2022traffic}
\bibinfo{author}{Yasin \surnamestart Bayzidi\surnameend}, \bibinfo{author}{Alen
  \surnamestart Smajic\surnameend}, \bibinfo{author}{Fabian \surnamestart
  H{\"u}ger\surnameend}, \bibinfo{author}{Ruby \surnamestart
  Moritz\surnameend}, \bibinfo{author}{Serin \surnamestart
  Varghese\surnameend}, \bibinfo{author}{Peter \surnamestart
  Schlicht\surnameend} \& \bibinfo{author}{Alois \surnamestart
  Knoll\surnameend} (\bibinfo{year}{2022}): \emph{\bibinfo{title}{Traffic sign
  classifiers under physical world realistic sticker occlusions: A cross
  analysis study}}.
\newblock In: {\slshape \bibinfo{booktitle}{2022 IEEE Intelligent Vehicles
  Symposium (IV)}}, \bibinfo{organization}{IEEE}, pp.
  \bibinfo{pages}{644--650}, \doi{10.1109/CVPR.2017.634}.

\bibitemdeclare{inproceedings}{boloor2019simple}
\bibitem{boloor2019simple}
\bibinfo{author}{Adith \surnamestart Boloor\surnameend}, \bibinfo{author}{Xin
  \surnamestart He\surnameend}, \bibinfo{author}{Christopher \surnamestart
  Gill\surnameend}, \bibinfo{author}{Yevgeniy \surnamestart
  Vorobeychik\surnameend} \& \bibinfo{author}{Xuan \surnamestart
  Zhang\surnameend} (\bibinfo{year}{2019}): \emph{\bibinfo{title}{Simple
  physical adversarial examples against end-to-end autonomous driving models}}.
\newblock In: {\slshape \bibinfo{booktitle}{2019 IEEE International Conference
  on Embedded Software and Systems (ICESS)}}, \bibinfo{organization}{IEEE}, pp.
  \bibinfo{pages}{1--7}, \doi{10.1145/1081870.1081950}.

\bibitemdeclare{misc}{sign}
\bibitem{sign}
\bibinfo{author}{Zahra \surnamestart Chaghazardi\surnameend}
  (\bibinfo{year}{2022}): \emph{\bibinfo{title}{Traffic Sign Detection using
  ILP}}.
\newblock
  \bibinfo{howpublished}{https://github.com/Chaghazardi/Traffic-Sign-Detection-using-ILP}.
\newblock
  \urlprefix\url{https://github.com/Chaghazardi/Traffic-Sign-Detection-using-ILP}.

\bibitemdeclare{inproceedings}{chaghazardi2023logic}
\bibitem{chaghazardi2023logic}
\bibinfo{author}{Zahra \surnamestart Chaghazardi\surnameend},
  \bibinfo{author}{Saber \surnamestart Fallah\surnameend} \&
  \bibinfo{author}{Alireza \surnamestart Tamaddoni-Nezhad\surnameend}
  (\bibinfo{year}{2023}): \emph{\bibinfo{title}{A Logic-based Compositional
  Generalisation Approach for Robust Traffic Sign Detection}}.
\newblock In: {\slshape \bibinfo{booktitle}{International Joint Conference on
  Artificial Intelligence 2023 Workshop on Knowledge-Based Compositional
  Generalization}}.

\bibitemdeclare{article}{chen2017attacking}
\bibitem{chen2017attacking}
\bibinfo{author}{Hongge \surnamestart Chen\surnameend}, \bibinfo{author}{Huan
  \surnamestart Zhang\surnameend}, \bibinfo{author}{Pin-Yu \surnamestart
  Chen\surnameend}, \bibinfo{author}{Jinfeng \surnamestart Yi\surnameend} \&
  \bibinfo{author}{Cho-Jui \surnamestart Hsieh\surnameend}
  (\bibinfo{year}{2017}): \emph{\bibinfo{title}{Attacking visual language
  grounding with adversarial examples: A case study on neural image
  captioning}}.
\newblock {\slshape \bibinfo{journal}{arXiv preprint arXiv:1712.02051}}.

\bibitemdeclare{inproceedings}{cropper2020turning}
\bibitem{cropper2020turning}
\bibinfo{author}{Andrew \surnamestart Cropper\surnameend},
  \bibinfo{author}{Sebastijan \surnamestart Dumancic\surnameend} \&
  \bibinfo{author}{Stephen~H \surnamestart Muggleton\surnameend}
  (\bibinfo{year}{2020}): \emph{\bibinfo{title}{Turning 30: New Ideas in
  Inductive Logic Programming}}.
\newblock In: {\slshape \bibinfo{booktitle}{IJCAI}},
  \doi{10.24963/ijcai.2020/673}.

\bibitemdeclare{misc}{metagol}
\bibitem{metagol}
\bibinfo{author}{Andrew \surnamestart Cropper\surnameend} \&
  \bibinfo{author}{Stephen~H. \surnamestart Muggleton\surnameend}
  (\bibinfo{year}{2016}): \emph{\bibinfo{title}{Metagol System}}.
\newblock \bibinfo{howpublished}{https://github.com/metagol/metagol}.
\newblock \urlprefix\url{https://github.com/metagol/metagol}.

\bibitemdeclare{inproceedings}{dai2018logical}
\bibitem{dai2018logical}
\bibinfo{author}{Wang-Zhou \surnamestart Dai\surnameend},
  \bibinfo{author}{Stephen \surnamestart Muggleton\surnameend},
  \bibinfo{author}{Jing \surnamestart Wen\surnameend}, \bibinfo{author}{Alireza
  \surnamestart Tamaddoni-Nezhad\surnameend} \& \bibinfo{author}{Zhi-Hua
  \surnamestart Zhou\surnameend} (\bibinfo{year}{2018}):
  \emph{\bibinfo{title}{Logical vision: One-shot meta-interpretive learning
  from real images}}.
\newblock In: {\slshape \bibinfo{booktitle}{Inductive Logic Programming: 27th
  International Conference, ILP 2017, Orl{\'e}ans, France, September 4-6, 2017,
  Revised Selected Papers 27}}, \bibinfo{organization}{Springer}, pp.
  \bibinfo{pages}{46--62}, \doi{10.1016/j.cviu.2007.08.003}.

\bibitemdeclare{inproceedings}{dai2015logical}
\bibitem{dai2015logical}
\bibinfo{author}{Wang-Zhou \surnamestart Dai\surnameend},
  \bibinfo{author}{Stephen~H \surnamestart Muggleton\surnameend} \&
  \bibinfo{author}{Zhi-Hua \surnamestart Zhou\surnameend}
  (\bibinfo{year}{2015}): \emph{\bibinfo{title}{Logical Vision:
  Meta-Interpretive Learning for Simple Geometrical Concepts.}}
\newblock In: {\slshape \bibinfo{booktitle}{ILP (Late Breaking Papers)}}, pp.
  \bibinfo{pages}{1--16}.

\bibitemdeclare{inproceedings}{deng2020analysis}
\bibitem{deng2020analysis}
\bibinfo{author}{Yao \surnamestart Deng\surnameend},
  \bibinfo{author}{Xi~\surnamestart Zheng\surnameend}, \bibinfo{author}{Tianyi
  \surnamestart Zhang\surnameend}, \bibinfo{author}{Chen \surnamestart
  Chen\surnameend}, \bibinfo{author}{Guannan \surnamestart Lou\surnameend} \&
  \bibinfo{author}{Miryung \surnamestart Kim\surnameend}
  (\bibinfo{year}{2020}): \emph{\bibinfo{title}{An analysis of adversarial
  attacks and defenses on autonomous driving models}}.
\newblock In: {\slshape \bibinfo{booktitle}{2020 IEEE international conference
  on pervasive computing and communications (PerCom)}},
  \bibinfo{organization}{IEEE}, pp. \bibinfo{pages}{1--10},
  \doi{10.1109/PerCom45495.2020.9127389}.

\bibitemdeclare{inproceedings}{duan2020adversarial}
\bibitem{duan2020adversarial}
\bibinfo{author}{Ranjie \surnamestart Duan\surnameend},
  \bibinfo{author}{Xingjun \surnamestart Ma\surnameend}, \bibinfo{author}{Yisen
  \surnamestart Wang\surnameend}, \bibinfo{author}{James \surnamestart
  Bailey\surnameend}, \bibinfo{author}{A~Kai \surnamestart Qin\surnameend} \&
  \bibinfo{author}{Yun \surnamestart Yang\surnameend} (\bibinfo{year}{2020}):
  \emph{\bibinfo{title}{Adversarial camouflage: Hiding physical-world attacks
  with natural styles}}.
\newblock In: {\slshape \bibinfo{booktitle}{Proceedings of the IEEE/CVF
  conference on computer vision and pattern recognition}}, pp.
  \bibinfo{pages}{1000--1008}.

\bibitemdeclare{article}{eykholt2018physical}
\bibitem{eykholt2018physical}
\bibinfo{author}{Kevin \surnamestart Eykholt\surnameend}, \bibinfo{author}{Ivan
  \surnamestart Evtimov\surnameend}, \bibinfo{author}{Earlence \surnamestart
  Fernandes\surnameend}, \bibinfo{author}{Bo~\surnamestart Li\surnameend},
  \bibinfo{author}{Amir \surnamestart Rahmati\surnameend},
  \bibinfo{author}{Florian \surnamestart Tramer\surnameend},
  \bibinfo{author}{Atul \surnamestart Prakash\surnameend},
  \bibinfo{author}{Tadayoshi \surnamestart Kohno\surnameend} \&
  \bibinfo{author}{Dawn \surnamestart Song\surnameend} (\bibinfo{year}{2018}):
  \emph{\bibinfo{title}{Physical adversarial examples for object detectors}}.
\newblock {\slshape \bibinfo{journal}{arXiv preprint arXiv:1807.07769}}
  \bibinfo{volume}{1}(\bibinfo{number}{3}), p.~\bibinfo{pages}{4}.

\bibitemdeclare{inproceedings}{eykholt2018robust}
\bibitem{eykholt2018robust}
\bibinfo{author}{Kevin \surnamestart Eykholt\surnameend}, \bibinfo{author}{Ivan
  \surnamestart Evtimov\surnameend}, \bibinfo{author}{Earlence \surnamestart
  Fernandes\surnameend}, \bibinfo{author}{Bo~\surnamestart Li\surnameend},
  \bibinfo{author}{Amir \surnamestart Rahmati\surnameend},
  \bibinfo{author}{Chaowei \surnamestart Xiao\surnameend},
  \bibinfo{author}{Atul \surnamestart Prakash\surnameend},
  \bibinfo{author}{Tadayoshi \surnamestart Kohno\surnameend} \&
  \bibinfo{author}{Dawn \surnamestart Song\surnameend} (\bibinfo{year}{2018}):
  \emph{\bibinfo{title}{Robust physical-world attacks on deep learning visual
  classification}}.
\newblock In: {\slshape \bibinfo{booktitle}{Proceedings of the IEEE conference
  on computer vision and pattern recognition}}, pp.
  \bibinfo{pages}{1625--1634}.

\bibitemdeclare{article}{farid2014plane}
\bibitem{farid2014plane}
\bibinfo{author}{Reza \surnamestart Farid\surnameend} \&
  \bibinfo{author}{Claude \surnamestart Sammut\surnameend}
  (\bibinfo{year}{2014}): \emph{\bibinfo{title}{Plane-based object
  categorisation using relational learning}}.
\newblock {\slshape \bibinfo{journal}{Machine Learning}} \bibinfo{volume}{94},
  pp. \bibinfo{pages}{3--23}, \doi{10.1007/s10994-013-5352-9}.

\bibitemdeclare{article}{gu2019badnets}
\bibitem{gu2019badnets}
\bibinfo{author}{Tianyu \surnamestart Gu\surnameend}, \bibinfo{author}{Kang
  \surnamestart Liu\surnameend}, \bibinfo{author}{Brendan \surnamestart
  Dolan-Gavitt\surnameend} \& \bibinfo{author}{Siddharth \surnamestart
  Garg\surnameend} (\bibinfo{year}{2019}): \emph{\bibinfo{title}{Badnets:
  Evaluating backdooring attacks on deep neural networks}}.
\newblock {\slshape \bibinfo{journal}{IEEE Access}} \bibinfo{volume}{7}, pp.
  \bibinfo{pages}{47230--47244}, \doi{10.1109/TKDE.2009.191}.

\bibitemdeclare{article}{gui2021review}
\bibitem{gui2021review}
\bibinfo{author}{Jie \surnamestart Gui\surnameend}, \bibinfo{author}{Zhenan
  \surnamestart Sun\surnameend}, \bibinfo{author}{Yonggang \surnamestart
  Wen\surnameend}, \bibinfo{author}{Dacheng \surnamestart Tao\surnameend} \&
  \bibinfo{author}{Jieping \surnamestart Ye\surnameend} (\bibinfo{year}{2021}):
  \emph{\bibinfo{title}{A review on generative adversarial networks:
  Algorithms, theory, and applications}}.
\newblock {\slshape \bibinfo{journal}{IEEE transactions on knowledge and data
  engineering}}.

\bibitemdeclare{inproceedings}{hendrik2017universal}
\bibitem{hendrik2017universal}
\bibinfo{author}{Jan \surnamestart Hendrik~Metzen\surnameend},
  \bibinfo{author}{Mummadi \surnamestart Chaithanya~Kumar\surnameend},
  \bibinfo{author}{Thomas \surnamestart Brox\surnameend} \&
  \bibinfo{author}{Volker \surnamestart Fischer\surnameend}
  (\bibinfo{year}{2017}): \emph{\bibinfo{title}{Universal adversarial
  perturbations against semantic image segmentation}}.
\newblock In: {\slshape \bibinfo{booktitle}{Proceedings of the IEEE
  international conference on computer vision}}, pp.
  \bibinfo{pages}{2755--2764}.

\bibitemdeclare{inproceedings}{kop2020machine}
\bibitem{kop2020machine}
\bibinfo{author}{Mauritz \surnamestart Kop\surnameend} (\bibinfo{year}{2020}):
  \emph{\bibinfo{title}{Machine learning \& EU data sharing practices}}.
\newblock In: {\slshape \bibinfo{booktitle}{TTLF Newsletter on Transatlantic
  Antitrust and IPR Developments}}, \bibinfo{organization}{Stanford-Vienna
  Transatlantic Technology Law Forum, Transatlantic Antitrust~…}.

\bibitemdeclare{incollection}{kurakin2018adversarial}
\bibitem{kurakin2018adversarial}
\bibinfo{author}{Alexey \surnamestart Kurakin\surnameend},
  \bibinfo{author}{Ian~J \surnamestart Goodfellow\surnameend} \&
  \bibinfo{author}{Samy \surnamestart Bengio\surnameend}
  (\bibinfo{year}{2018}): \emph{\bibinfo{title}{Adversarial examples in the
  physical world}}.
\newblock In: {\slshape \bibinfo{booktitle}{Artificial intelligence safety and
  security}}, \bibinfo{publisher}{Chapman and Hall/CRC}, pp.
  \bibinfo{pages}{99--112}, \doi{10.1201/9781351251389-8}.

\bibitemdeclare{misc}{uber}
\bibitem{uber}
\bibinfo{author}{T.~B. \surnamestart Lee\surnameend} (\bibinfo{year}{2018}):
  \emph{\bibinfo{title}{Software bug led to death in ubers self-driving
  crash}}.
\newblock
  \urlprefix\url{https://arstechnica.com/tech-policy/2018/05/report-software-bug-led-to-death-in-ubers-self-driving-crash/}.

\bibitemdeclare{article}{li2020adaptive}
\bibitem{li2020adaptive}
\bibinfo{author}{Yujie \surnamestart Li\surnameend}, \bibinfo{author}{Xing
  \surnamestart Xu\surnameend}, \bibinfo{author}{Jinhui \surnamestart
  Xiao\surnameend}, \bibinfo{author}{Siyuan \surnamestart Li\surnameend} \&
  \bibinfo{author}{Heng~Tao \surnamestart Shen\surnameend}
  (\bibinfo{year}{2020}): \emph{\bibinfo{title}{Adaptive square attack: Fooling
  autonomous cars with adversarial traffic signs}}.
\newblock {\slshape \bibinfo{journal}{IEEE Internet of Things Journal}}
  \bibinfo{volume}{8}(\bibinfo{number}{8}), pp. \bibinfo{pages}{6337--6347},
  \doi{10.1109/CVPR.2018.00957}.

\bibitemdeclare{misc}{morgulis2019fooling}
\bibitem{morgulis2019fooling}
\bibinfo{author}{Nir \surnamestart Morgulis\surnameend},
  \bibinfo{author}{Alexander \surnamestart Kreines\surnameend},
  \bibinfo{author}{Shachar \surnamestart Mendelowitz\surnameend} \&
  \bibinfo{author}{Yuval \surnamestart Weisglass\surnameend}
  (\bibinfo{year}{2019}): \emph{\bibinfo{title}{Fooling a real car with
  adversarial traffic signs}}.
\newblock \urlprefix\url{https://arxiv.org/abs/1907.00374}.

\bibitemdeclare{article}{muggleton1991inductive}
\bibitem{muggleton1991inductive}
\bibinfo{author}{Stephen \surnamestart Muggleton\surnameend}
  (\bibinfo{year}{1991}): \emph{\bibinfo{title}{Inductive logic programming}}.
\newblock {\slshape \bibinfo{journal}{New generation computing}}
  \bibinfo{volume}{8}(\bibinfo{number}{4}), pp. \bibinfo{pages}{295--318},
  \doi{10.1007/BF03037089}.

\bibitemdeclare{article}{muggleton2015meta}
\bibitem{muggleton2015meta}
\bibinfo{author}{Stephen~H \surnamestart Muggleton\surnameend},
  \bibinfo{author}{Dianhuan \surnamestart Lin\surnameend} \&
  \bibinfo{author}{Alireza \surnamestart Tamaddoni-Nezhad\surnameend}
  (\bibinfo{year}{2015}): \emph{\bibinfo{title}{Meta-interpretive learning of
  higher-order dyadic datalog: Predicate invention revisited}}.
\newblock {\slshape \bibinfo{journal}{Machine Learning}}
  \bibinfo{volume}{100}(\bibinfo{number}{1}), pp. \bibinfo{pages}{49--73},
  \doi{10.1007/s10994-014-5471-y}.

\bibitemdeclare{article}{salami2021state}
\bibitem{salami2021state}
\bibinfo{author}{Bukola \surnamestart Salami\surnameend},
  \bibinfo{author}{Keijo \surnamestart Haataja\surnameend} \&
  \bibinfo{author}{Pekka \surnamestart Toivanen\surnameend}
  (\bibinfo{year}{2021}): \emph{\bibinfo{title}{State-of-the-Art Techniques in
  Artificial Intelligence for Continual Learning: A Review.}}
\newblock {\slshape \bibinfo{journal}{FedCSIS (Position Papers)}}, pp.
  \bibinfo{pages}{23--32}.

\bibitemdeclare{article}{shanahan2005perception}
\bibitem{shanahan2005perception}
\bibinfo{author}{Murray \surnamestart Shanahan\surnameend}
  (\bibinfo{year}{2005}): \emph{\bibinfo{title}{Perception as abduction:
  Turning sensor data into meaningful representation}}.
\newblock {\slshape \bibinfo{journal}{Cognitive science}}
  \bibinfo{volume}{29}(\bibinfo{number}{1}), pp. \bibinfo{pages}{103--134},
  \doi{10.1207/s15516709cog2901_5}.

\bibitemdeclare{article}{stallkamp2012man}
\bibitem{stallkamp2012man}
\bibinfo{author}{Johannes \surnamestart Stallkamp\surnameend},
  \bibinfo{author}{Marc \surnamestart Schlipsing\surnameend},
  \bibinfo{author}{Jan \surnamestart Salmen\surnameend} \&
  \bibinfo{author}{Christian \surnamestart Igel\surnameend}
  (\bibinfo{year}{2012}): \emph{\bibinfo{title}{Man vs. computer: Benchmarking
  machine learning algorithms for traffic sign recognition}}.
\newblock {\slshape \bibinfo{journal}{Neural networks}} \bibinfo{volume}{32},
  pp. \bibinfo{pages}{323--332}, \doi{10.1016/j.neunet.2012.02.016}.

\bibitemdeclare{article}{szegedy2013intriguing}
\bibitem{szegedy2013intriguing}
\bibinfo{author}{Christian \surnamestart Szegedy\surnameend},
  \bibinfo{author}{Wojciech \surnamestart Zaremba\surnameend},
  \bibinfo{author}{Ilya \surnamestart Sutskever\surnameend},
  \bibinfo{author}{Joan \surnamestart Bruna\surnameend},
  \bibinfo{author}{Dumitru \surnamestart Erhan\surnameend},
  \bibinfo{author}{Ian \surnamestart Goodfellow\surnameend} \&
  \bibinfo{author}{Rob \surnamestart Fergus\surnameend} (\bibinfo{year}{2013}):
  \emph{\bibinfo{title}{Intriguing properties of neural networks}}.
\newblock {\slshape \bibinfo{journal}{arXiv preprint arXiv:1312.6199}}.

\bibitemdeclare{inproceedings}{thys2019fooling}
\bibitem{thys2019fooling}
\bibinfo{author}{Simen \surnamestart Thys\surnameend}, \bibinfo{author}{Wiebe
  \surnamestart Van~Ranst\surnameend} \& \bibinfo{author}{Toon \surnamestart
  Goedem{\'e}\surnameend} (\bibinfo{year}{2019}): \emph{\bibinfo{title}{Fooling
  automated surveillance cameras: adversarial patches to attack person
  detection}}.
\newblock In: {\slshape \bibinfo{booktitle}{Proceedings of the IEEE/CVF
  conference on computer vision and pattern recognition workshops}}, pp.
  \bibinfo{pages}{0--0}.

\bibitemdeclare{misc}{pyilp}
\bibitem{pyilp}
\bibinfo{author}{Dany \surnamestart Varghese\surnameend}
  (\bibinfo{year}{2022}): \emph{\bibinfo{title}{PyILP}}.
\newblock \bibinfo{howpublished}{https://github.com/danyvarghese/PyILP}.
\newblock \urlprefix\url{https://github.com/danyvarghese/PyILP}.

\bibitemdeclare{misc}{Yadav}
\bibitem{Yadav}
\bibinfo{author}{\surnamestart {Vivek Yadav}\surnameend}
  (\bibinfo{year}{2016}): \emph{\bibinfo{title}{German sign classification
  using deep learning neural networks}}.
\newblock
  \bibinfo{howpublished}{\url{https://github.com/vxy10/p2-TrafficSigns}}.

\bibitemdeclare{article}{yan2022rolling}
\bibitem{yan2022rolling}
\bibinfo{author}{Chen \surnamestart Yan\surnameend}, \bibinfo{author}{Zhijian
  \surnamestart Xu\surnameend}, \bibinfo{author}{Zhanyuan \surnamestart
  Yin\surnameend}, \bibinfo{author}{Xiaoyu \surnamestart Ji\surnameend} \&
  \bibinfo{author}{Wenyuan \surnamestart Xu\surnameend} (\bibinfo{year}{2022}):
  \emph{\bibinfo{title}{Rolling Colors: Adversarial Laser Exploits against
  Traffic Light Recognition}}.
\newblock {\slshape \bibinfo{journal}{arXiv preprint arXiv:2204.02675}}.

\end{thebibliography}
\end{document}